# Segmentation method of U-net sheet metal engineering drawing based on CBAM attention mechanism

Songzhiwei[1], Yao Hui[2]

[1] songzhiwei@st.xatu.edu.cn；[2] yaohui@xatu.edu.cn

**Abstract**

In the manufacturing process of heavy industrial equipment, the specific unit in the welding diagram is first manually redrawn and then the corresponding sheet metal parts are cut, which is inefficient. To this end, this paper proposes a U-net-based method for the segmentation and extraction of specific units in welding engineering drawings. This method enables the cutting device to automatically segment specific graphic units according to visual information and automatically cut out sheet metal parts of corresponding shapes according to the segmentation results. This process is more efficient than traditional human-assisted cutting. Two weaknesses in the U-net network will lead to a decrease in segmentation performance: first, the focus on global semantic feature information is weak, and second, there is a large dimensional difference between shallow encoder features and deep decoder features. Based on the CBAM (Convolutional Block Attention Module) attention mechanism, this paper proposes a U-net jump structure model with an attention mechanism to improve the network's global semantic feature extraction ability. In addition, a U-net attention mechanism model with dual pooling convolution fusion is designed, the deep encoder's maximum pooling + convolution features and the shallow encoder's average pooling + convolution features are fused vertically to reduce the dimension difference between the shallow encoder and deep decoder. The dual-pool convolutional attention jump structure replaces the traditional U-net jump structure, which can effectively improve the specific unit segmentation performance of the welding engineering drawing. Using vgg16 as the backbone network, experiments have verified that the IoU, mAP, and Accu of our model in the welding engineering drawing dataset segmentation task are 84.72%, 86.84%, and 99.42%, respectively. The method in this paper is 22.10, 19.09, and 0.05 percentage points higher than the traditional U-net, and has better segmentation performance for welding engineering drawings.

## 1. Introduction

Heavy industry equipment generally adopts customized manufacturing. In such projects,

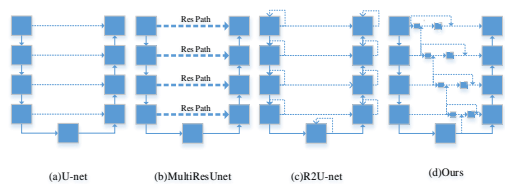

Fig 1. Comparison of our method (d) with skip structure schemes of other models. Dashed lines denote skip connections.

large sheet metal parts need to be cut out according to the content of customized engineering drawings and finally manufactured by welding and stamping operations. Efficiency and precision are critical to the manufacture of customized heavy industrial equipment. The specific process of the traditional way to obtain sheet metal parts in the manufacturing process of heavy industry equipment is manual recognition of engineering drawings[1,2]-manual redrawing of specific graphics[3,4]-sheet metal cutting and stamping based on CAD/CAM integrated system[5,6,7]. The process of obtaining specific sheet metal parts in traditional ways is cumbersome and inefficient.

With the concept of intelligent manufacturing being proposed, the efficient automatic cutting of customized sheet metal parts is an urgent problem to be solved in the current heavy industry equipment manufacturing process. In recent years, with the development of artificial intelligence, the industrial field is also more inclined to use deep learning methods to solve engineering problems. Some studies have shown that this method has an excellent performance in solving engineering problems. For example, Lau et al[8]. used deep learning image segmentation technology to realize the automatic detection of road cracks, and achieved good performance by using the ResNet-34 pre-training model. Kazemifar et al[9]. used the 2DU-net segmentation model to diagnose lesions in male pelvic CT, and achieved high-precision automatic segmentation of ROI. Hou et al[10]. applied the deep learning method to welding engineering to realize the detection of welding defects. According to the high precision and low complexity of the U-net[11] model for image segmentation tasks, this paper proposes an improved U-net model based on deep learning methods for the segmentation of specific units in sheet metal engineering drawings in heavy industrial equipment manufacturing. This method is mainly used to solve the problem of low efficiency of obtaining sheet metal parts by traditional methods. The overall structure of the U-net model is composed of an encoder and a decoder, and it uses a jump-connected structure to achieve the fusion of global semantic features. In the process of image semantic segmentation, the encoder encodes the extracted low-dimensional image features, and the decoder decodes the encoded semantic features. Through the jump structure, the low-dimensional features after the convolution of the encoder and the high-dimensional features after the upsampling of the decoder can be fused, eliminating the loss of some image features in the encoding and decoding process, and achieving better global segmentation. This paper uses the traditional U-net to conduct a feasibility experiment on the task of sheet metal welding graphics segmentation, and further studies and analyzes the structure and principle of U-Net by using the attention mechanism[12,13,14]. Finally, it is proposed to add a CBAM attention mechanism to improve the model's ability to extract global semantic features, and an improved U-net model with a dual-pool convolution fusion attention mechanism is designed. This model can better realize the global feature information fusion of encoder and decoder features, and effectively reduce the dimension difference between low-dimensional encoder and high-dimensional decoder in the process of feature fusion. It can better realize the segmentation of specific units in sheet metal engineering drawings. The main contributions of this study are as follows:

- Propose the automatic cutting technology of sheet metal parts based on U-net for heavy industry equipment manufacturing.
- The segmentation method combining CBAM and U-net is proposed to be suitable for (non-human-assisted) high-precision segmentation of sheet metal graphics.
- A double pooling + convolutional skip structure is proposed to reduce the dimensional difference between encoding and decoding features.
- A skip structure of vertical coding features is proposed to improve the fusion of global semantic information.
- The improved model has been verified for its high-performance segmentation capability of sheet metal graphics.

## 2. Related Work

According to the successful application of deep learning image segmentation technology in many engineering problems, this paper proposes a U-net sheet metal engineering drawing segmentation method based on the CBAM attention mechanism. As we all know, many excellent segmentation networks already exist in deep learning image segmentation. The fully convolutional segmentation model (Fully Convolutional Networks, FCNs) of deep learning was first proposed by Jonathan Long [15]. FCNs abandon the traditional fully connected layer, and the overall network structure uses a fully convolutional method to achieve end-to-end pixel-level dense prediction of image features, which is suitable for more complex global semantic feature segmentation tasks. The fully convolutional segmentation model uses transposed convolution operations to obtain semantically segmented images of the original size through upsampling. The input of FCNs is an RGB image of any size and the output is the same size as the input. At the same time, it proposes a classic skip connection for fusing features from deep layers (including classification information) and intermediate layers (including location information) to improve feature accuracy output. U-net can be considered a variation of FCN, which still uses the encoder-decoder and skip structure. Compared with the former, the unique skip connection architecture of the U-net enables the decoder to obtain more spatial information lost during the pooling operation and restore a complete spatial resolution. The semantic difference between the encoder and decoder is reduced to achieve better segmentation performance. U-Net mainly has two cores: (1) Dimensional difference problem between the low-dimensional encoder and high-dimensional decoder in the process of semantic information fusion. How to effectively reduce the dimension difference in the image fusion process of the encoder and decoder? (2) The problem of image spatial position information, how can the encoder and decoder realize the learning of image spatial position information? Researchers have introduced many methods to solve the above problems to reduce the incompatible feature differences between these two groups.

Whether learning low-resolution Deeplab-v1, Deeplab-v2, Deeplab-v3, PSPNet[16,17,18,19] or recovering high-resolution SegNet, DeconvNet[20,21] and parallel high and low-resolution HRNet[22]. Even deep neural networks ENet and UpperNet can achieve real-time prediction[23,24]. The segmentation networks focus on acquiring global image semantic features and spatial location information. Therefore, the model mainly integrates the global upper and lower semantic features in image segmentation and learns spatial location feature information. In recent years, there have been many improved models based on U-net. For example, R2UNet, R2U++Net, CAggNet, MultiResUNet, NonlocalUNets and UCTransNet[25,26,27,28,29,30], etc. These networks are all improvements to the U-net skip structure to achieve a better fusion of global contextual feature information in the encoder and decoder and achieve excellent segmentation performance, A network that improves segmentation performance by improving the U-net jump structure, as shown in Fig 1. The U-net skip structure can better realize the fusion of global semantic information. However, as mentioned in UNet++[31], the skip-structure front-end encoder semantic features have lower dimensionality than the back-end decoder semantic features. Therefore, there is a large difference in feature dimension in the skip structure, which makes the segmentation performance not good enough.

On the other hand, Fei Wang[12] proposed a residual attention network using an encoder-decoder approach. Based on this,

### 2.1 CBAM-U-net Model

The Convolutional Block Attention Module(CBAM)[13] can improve the ability of the convolutional network to express the image of the feature layer, in the process of extracting image features, pay more attention to the feature factors that have a greater impact on the target, and suppress the expression of non-important features that have no obvious impact. Input the original sheet metal engineering drawing F, which is transformed into a featured image X through pooling + two convolution operations.

Sanghyun Woo[13] proposed the CBAM (Convolutional Block Attention Module) module to realize the entire convolution channel semantic and spatial information calculation. UCTransNet is a recently proposed attention module inspired by the Self Attention Mechanism and Multi-Head Attention mechanisms in Transformer[32], and its purpose is to enable the encoder-decoder to obtain more global information fusion. Based on the above studies, this paper proposes a U-net model based on the CBAM attention mechanism, which considers the information fusion of channel and spatial dimensions and the dimensional differences between encoder and decoder features. Besides, This article proposes global information linkage between U-net encoders, which includes feature cluster integration between vertical encoders and vertical and horizontal double-pooling convolutions. The output features are fused with the original features through the attention module, and the whole jump to the high-dimensional feature layer of the decoder to achieve secondary fusion. The improved U-net model can not only better extract global semantic features but also reduce the semantic differences in the process of encoder-decoder feature fusion. Through the experimental verification, the method in this paper has better performance in the sheet metal graphics segmentation task.

The shallow feature map $X \in \mathbb{R}^{C \times H \times W}$ is input to the CBAM module, which infers the channel attention map Tc(X) and the planar spatial attention map Ts(X'), as shown in Fig 2. The overall attention process of the module is rough as follows:

$$X' = Tc(X) \otimes X \qquad (1.1)$$

$$X'' = Ts(X') \otimes X' \qquad (1.2)$$

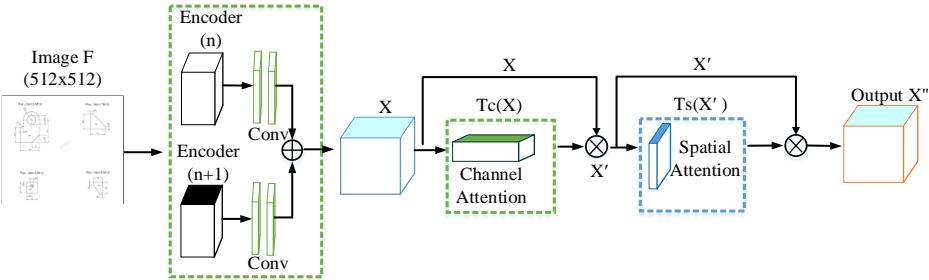

Fig 2. CBAM architecture. This module comprises a channel module and a spatial attention module consecutively. The encoder feeds the double-pooled and convolutional features into this module, and the CBAM generates global features with channel and spatial location information. (n) and (n+1) respectively represent the encoders of different vertical layers of U-net.

Where ⊗ means element-wise multiplication. In the operation process, the CBAM module can continuously obtain the 1D channel attention feature map $Tc(X) \in \mathbb{R}^{C \times 1 \times 1}$ and the 2D spatial attention feature map $Ts(X') \in \mathbb{R}^{1 \times H \times W}$ according to the input feature map $X \in \mathbb{R}^{C \times H \times W}$. In image feature extraction, the CBAM module assigns corresponding attention weights to the image features that have a greater influence on the target task (the attention weights in the spatial direction are propagated in the channel direction, and vice versa). The feature map X'' marked with channel-spatial attention weights calculated and output by the CBAM module is finer than the image features output by traditional U-net. At the same time, X'' is upsampled by 2x2 and fused with the low-dimensional feature cluster of the horizontal encoder, and the final feature F' is output. Image features F' are combined with decoder upsampled graph feature clusters to reduce global contextual semantic differences. The overall improved U-net network structure is shown in Fig 3.

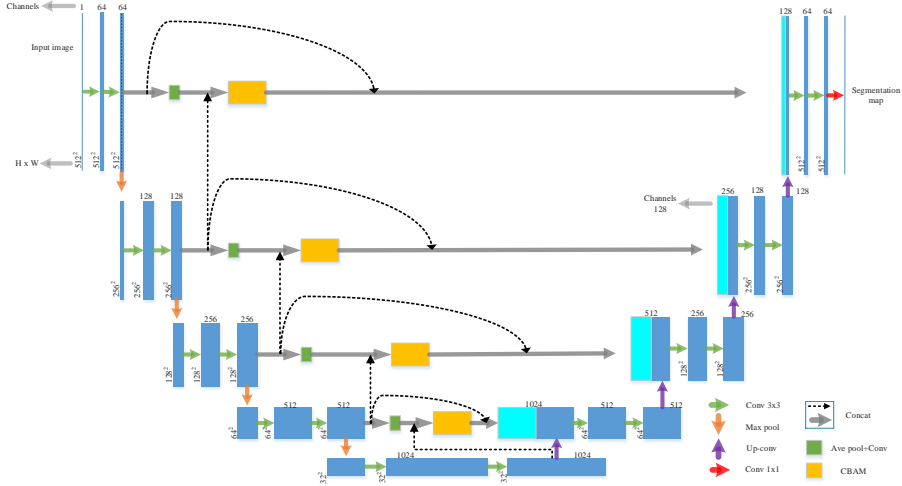

Fig 3. The improved model overall architecture is proposed in this paper (input raw image pixel 512x512). Green squares represent average pooling and two convolution operations, Orange squares represent the CBAM attention mechanism module, and indigo squares represent feature cluster integration. Different colored arrows indicate different operations.

## 2.2 U-net Attention Module

The residual attention network[12] adopts pre-activated residual units ResNeXt[33] and Inception[34] as a two-branch parallel network structure stacked with residual attention modules. Bolei Zhou and Jie Hu et al[35,36] used average pooling to aggregate and count spatial information, respectively. The convolutional attention module uses inter-channel feature relationships to compress the spatial dimension of the input feature map to compute channel attention. Moreover, it is verified that the average pooling and max pooling simultaneously can improve the feature network's representation ability. What is mentioned here is that the traditional CBAM directly performs the maximum pooling and average pooling operations on the image input. CBAM uses average pooling and maximum pooling to fuse the spatial information of semantic features to generate two different global semantic space information expression features $F^c_{avg}$ and $F^c_{max}$. In improving the U-net model structure, we made a small change to this. Instead of directly inputting the sheet metal image F to the CBAM module, we input the features extracted and fused by the vertical encoder. Feed the fused features X into a shared multi-layer perceptron (MLP). Therefore, at this time, CBAM uses average pooling and maximum pooling operations to generate two different global semantic space information expression features: $X^c_{avg} \in \mathbb{R}^{C \times 1 \times 1}$ and $X^c_{max} \in \mathbb{R}^{C \times 1 \times 1}$. At the same time, the shared multi-layer perceptron performs integration + ReLU operation on the input feature clusters to generate channel attention feature maps $Tc(X) \in \mathbb{R}^{C \times 1 \times 1}$. $Tc(X)$ contains the attention feature relations between the various channel axes of the input feature X. Then, $Tc(X)$ is fused with the input feature X (Note: X is the image feature after the original image F has been pooled and convolved by the upper and lower encoders) to generate a feature map X'. According to the semantic difference between the Upper-encoder and Lower-encoder features, the convolution and linear rectification unit (ReLU) are used to continue to calculate the spatial information relationship of the feature X', and generate the attention space feature map

Ts(X'). The channel attention feature X' is merged with the spatial position attention feature Ts(X') to generate a globally informative feature X" with channel-spatial dual attention. The detailed calculation process of the attention module is as follows.

$$Tc(X) = S(M(Maxpool(X)) + M(Avgpool(X)))$$
$$= S\left(W_2\left(W_1\left(X^c_{avg}\right)\right) + W_2\left(W_1\left(X^c_{max}\right)\right)\right) \quad (1.3)$$

Where $Tc(X)$ is a 1-dimensional channel attention image feature, S is a sigmoid activation function, and M represents a multi-layer perceptron (MLP) shared layer. *Maxpool*(X) and *Avgpool*(X) are the secondary pooling operations of horizontal low-dimensional encoder features and vertical high-dimensional encoder features, respectively (X is the result of vertical maximum pooling and horizontal average pooling). $W_1$ and $W_2$ represent the shared weights of the input multi-layer perceptron (MLP). The output of the ReLU activation layer is $W_1$. The extraction process of spatial feature information in the CBAM structure is similar to that of channel feature extraction. The difference is that the sheet metal graphic feature F is input into the CBAM module as X after the convolution operation. The CBAM module still uses maximum pooling and minimum pooling to fuse image feature semantic information to generate two 2D images: $X^s_{avg} \in \mathbb{R}^{1 \times H \times W}$ and $X^s_{max} \in \mathbb{R}^{1 \times H \times W}$. And perform a convolution operation on it to generate an image Ts(X') containing spatial feature information. (It is worth noting that we use three 3x3 convolutions instead of 7x7 convolutions in CBAM to reduce calculation parameters). The image spatial feature information is calculated as follows:

$$Ts(X') = S\left(f_{3x3x3}(Tc([Avgpool(X); Maxpool(X)]))\right)$$
$$= S\left(f_{3x3x3}\left(\left[X^s_{avg}; X^s_{max}\right]\right)\right) \quad (1.4)$$

Ts(X') is a 2-dimensional spatial position information feature, and f3x3x3 represents three 3x3 convolution operations. Among them, the spatial attention feature layer performs average pooling and maximum pooling operations on the channel attention feature layer Tc(X) during the calculation process, instead of inputting the welding graph F. The detailed structure of the attention module is shown in Fig 4.

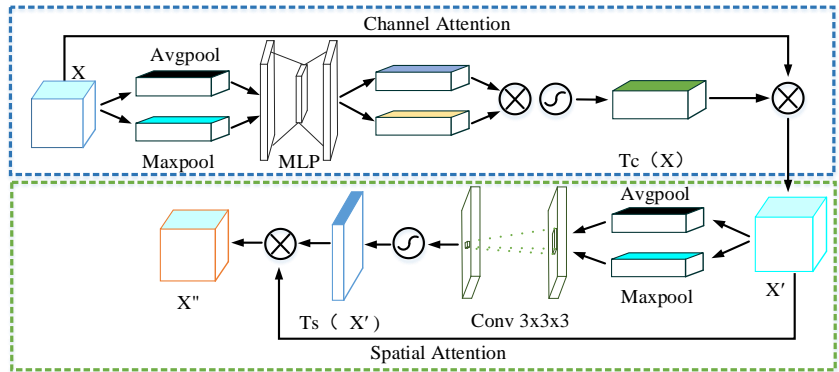

Fig 4. Principle of attention mechanism. X is used as input to the multi-layer perceptron (MLP), and the feature X' with channel attention information is generated through feature cluster multiplication and Softmax operation. X' output features X" with channel spatial information through a similar operation of the spatial attention module.

This study improves the U-net segmentation network, the main purpose of which is to use artificial intelligence to apply it to the manufacturing process of heavy industry equipment welding engineering, so that it can automatically cut the entire sheet metal, thereby liberating labor and improving enterprise efficiency. Encapsulate the improved model after training in an integrated cutting and processing center with visual functions to realize the automatic cutting of sheet metal parts in the process of intelligent manufacturing. The specific process is shown in Fig 5. At present, this research has been initially applied to the cutting processing center of China Metallurgical Equipment Corporation to realize a simple sheet metal cutting test.

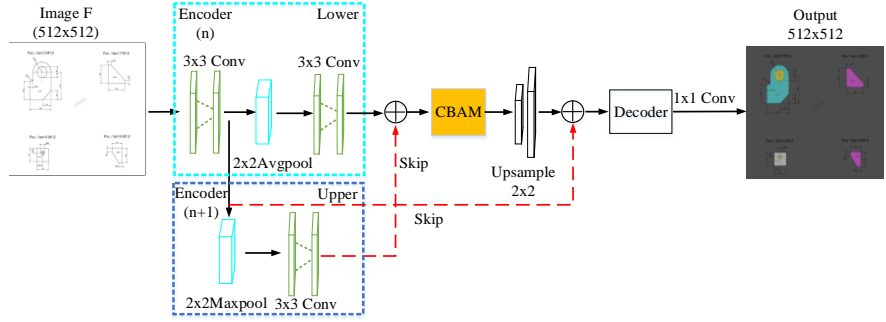

Fig 5. Improve the U-net model cutting sheet metal specific contour mechanism. Segmentation extracts the specific unit of welding engineering graphics, and the cutting device automatically cuts the corresponding parts on the whole sheet metal relying on vision.

## 3. Experiments

### 3.1 *Dataset*

The training data set is a non-public engineering atlas of complex welding structures used in the manufacture of heavy industry equipment provided by the cooperative heavy-pressure riveting and welding company, as well as some public welding engineering atlases in the United States and Japan. Its quantity is shown in Table 1. It is well known that the size of the dataset directly affects the training results, and the network may overfit when there are few training samples. To avoid the problem of biased training results due to the small number of data sets, data set enhancement processing is performed on the provided data sets. First of all, this study uses manual annotation to select 600 welding engineering graphics provided by China Metallurgical Group, the United States, and Japan from the welding equipment engineering atlas collection. The dataset was augmented by cropping, mirroring, deflecting, adding noise, etc., resulting in a dataset of 4094 annotated engineering drawings. Then, modify these datasets into training atlases in VOC format.

|  | MCC | America | Japan | Total |
|---|---|---|---|---|
| Training set | 1637 | 844 | 794 | 3275 |
| Validation set | 205 | 106 | 99 | 410 |
| Test set | 205 | 105 | 99 | 409 |
| Total number | 2047 | 1055 | 992 | 4094 |

Table 1. Sources of welding engineering datasets and the number of datasets after data enhancement processing.

### 3.2 *Evaluation metrics*

$$IoU = \frac{Area(Pp \cap Pgt)}{Area(Pp \cup Pgt)} \quad (1.5)$$

$Pp$ Is the prediction frame, $Pgt$ Is the ground truth frame, and $IoU$ is the intersection area of the $Pp$ $Pgt$ regions divided by the union area.

$$Accuracy = \frac{(TP + TN)}{(TP + FP + FN + TN)} \quad (1.6)$$

$TP$ Means that the sample is positive and the predicted value is positive.
$FP$ Means that the sample is negative and the predicted value is positive.
$FN$ Means that the sample is positive and the predicted value is negative.
$TN$ Means that the sample is negative and the predicted value is negative.

Accuracy is used to judge the accuracy of the prediction results, the total number of correct predictions/the total number of samples. There are two cases when the prediction is accurate: the sample is positive, the prediction is positive, and the sample is negative.

$$AP = \frac{\sum Precision}{N} \quad (1.7)$$

Precision is the sample precision, and *N* is the total number of samples.

$$mAP = \frac{\sum AP}{Nclass} \qquad (1.8)$$

$AP$ It is the average precision and $Nclass$ the number of classes. $AP$ The average precision rate is the sum of the precision rates for each

### 3.3 Implementation details

This experiment is carried out on the environment framework environment of Anaconda3, using the GPU accelerated training method of CUDA parallel computing architecture. Vgg16[37] is used as the backbone network of the overall structure, and its internal parameters are trained with the Adam optimizer of 0.9. The initial learning rate is set to 0.0001, and the learning rate drop method is cos drop. In addition, by combining the actual requirements and comparing the use of the Focal loss with better results, the main purpose is to reduce the weight of easily distinguishable samples and focus on samples that are difficult to distinguish. The experimental comparison results are shown in Table 2. The overall process trains for a total of 100 epochs. When the network is trained to the 50th epoch, the network starts to load and evaluate the validation set, and the network model fluctuates slightly during this process. To evaluate the model, the K-fold cross-validation method (K=5) was used to validate the model, as shown in Fig 6. The performance of the improved welded graph U-net segmentation model was evaluated by using the Intersection Over Union (IoU), Accuracy (Accu), and Mean Prediction Precision (mAP) metrics. Its training loss curve is shown in Fig 7. To directly demonstrate the segmentation effect of our proposed U-net improved model on welding engineering drawings, the results are compared with the traditional U-net segmentation effect through ablation experiments. Its visualization effect is shown in Fig 8. By comparing the traditional U-net segmentation performance with sample (of a particular category) divided by the total number of samples. $mAP$ Is the mean, and the average precision is the sum of the $AP$ values of all categories divided by the number of categories (note: $mAP$ in the table is the $mAP$ of the validation set).

the horizontal encoder average pooling + double convolution vertical encoder jump structure fusion method proposed in this paper, a comparative experiment is carried out. After that, we continue to compare the traditional U-net segmentation performance with the crossbar encoder jump structure of the attention mechanism CBAM proposed in this paper. The results of the comparative experiments are shown in Table 3. The segmentation performance of the improved model proposed in this paper when applied to welding engineering drawings is higher than that of traditional U-net.

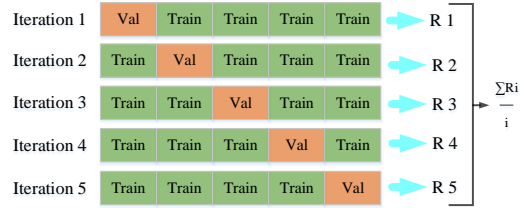

Fig 6. The welding engineering atlas adopts a K-fold cross-training verification process, the training set and verification sets are 4:1, and the stratification factor is K=5.

| Network | Loss function | Epoch | IoU | Accu |
|---|---|---|---|---|
| CBAM-U-net | CE Loss | 100 | 0.8252 | 0.9810 |
|  | BCE Loss | 100 | 0.8312 | 0.9793 |
|  | Poly Loss | 100 | 0.8400 | 0.9902 |
|  | Focal Loss | 100 | **0.8472** | **0.9942** |

Table 2. The performance comparison results of various loss functions used by CBAM-U-net to deal with imbalanced datasets.

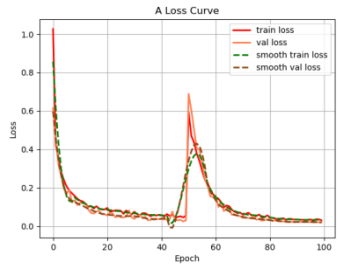 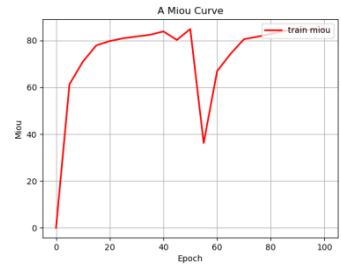

Fig 7. The loss curve graph during training and the 50th epoch model reaches a state of convergence. When training to 50 epochs, the network starts to unfreeze the evaluation model. The model will be reloaded from its original form, and fluctuations will have no effect.

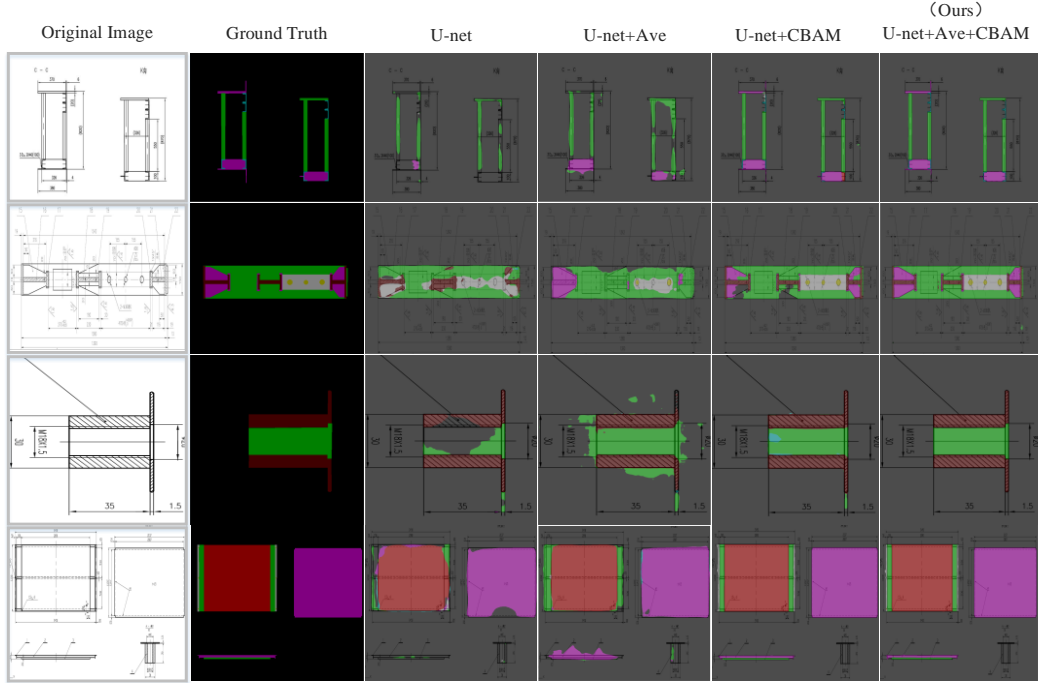

Fig 8. Visual comparison of segmentation effects between different methods. The original input is a welded structure drawing, and the second column is the ground truth mask. Where Ave is denoted as the average pooling and convolution operations as green squares in Fig 3, CBAM is the attention module, as shown in the orange court in Fig 3.

To avoid problems such as the increased variance of the estimated value and mean shift when the network model extracts features. In the U-net improved model, this paper proposes to use horizontal low-dimensional encoder average pooling and vertical high-dimensional encoder maximum pooling to better capture global feature information. After the encoder performs a double pooling operation, its horizontal encoder continues to perform two convolution operations to extract higher-dimensional spatial information of the image. The input image resolution is 512x512, and the encoder uses a repeated convolution operation of 3x3 (same padding), followed by a linear rectification unit (ReLU). The vertical high-dimensional encoder uses max-pooling with stride 2 of size 2x2, and the horizontal encoder uses equal-sized average pooling. Perform two 3x3 (same padding) convolution operations on the average pooled semantic features. The semantic feature clusters of the vertical high-dimensional encoder after max-pooling convolution are fused with the semantic feature clusters of the horizontal low-dimensional encoder after average pooling convolution. The fused feature clusters are input into the jump structure module of the CBAM attention mechanism to realize global context information fusion. The feature information output by the improved model added to the attention mechanism module is fused with the feature information of the low-dimensional encoder (this low-dimensional feature information does not perform any pooling operation). Finally, the dimensionality reduction of the fused semantic features is performed through 1x1 convolution through the jump structure, and the features with a resolution of 512x512 channels and 64 channels after upsampling by the decoder are fused again. Perform 3x3 convolution and 1x1 convolution on the integrated feature map to realize the mapping of each component feature vector class and achieve its segmentation effect.

| Method | IoU | mAP | Accu |
|---|---|---|---|
| Base (U-net) [11] | 0.6262 | 0.6775 | 0.9937 |
| Base +Ave | 0.6324 | 0.6947 | 0.9605 |
| Base+CBAM | 0.7299 | 0.7549 | 0.9745 |
| Base+Ave+CBAM | **0.8472** | **0.8684** | **0.9942** |

Table 3. Comparison of welding engineering map segmentation by different methods, Ave is denoted as average pooling and convolution operations, and CBAM is denoted as attention module. Same benchmark, bold font means excellent.

To further verify the scientificity and effectiveness of the method proposed in this paper, we use CNN to continue to use this method to segment and extract the specific outline of the welding engineering drawing. Abandoning the jumping structure of the traditional U-net, the method proposed in this paper is applied to the traditional convolutional neural network to perform segmentation and extraction experiments on welding graphics. The model uses consecutive 3x3 convolution operations, and the last layer uses the principle of mapping to achieve the segmentation effect. By using double-pooling convolution operations and adding attention mechanism operations in the CNN network model, different CNN models are compared for segmentation performance experiments. Fig 9 is the training loss curve of the model with double-pooling convolution fusion and attention mechanism added at the same time. The visualization result of the specific contour segmentation of the welding engineering drawing is shown in Fig 10.

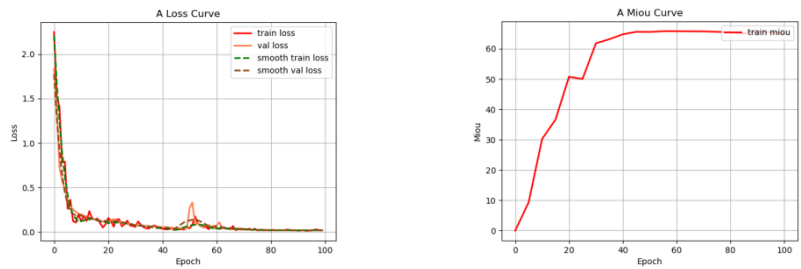

Fig 9. A graph of the loss curve of a continuous convolutional CNN. When the training has gone through 45 epochs, the model reaches the state of convergence.

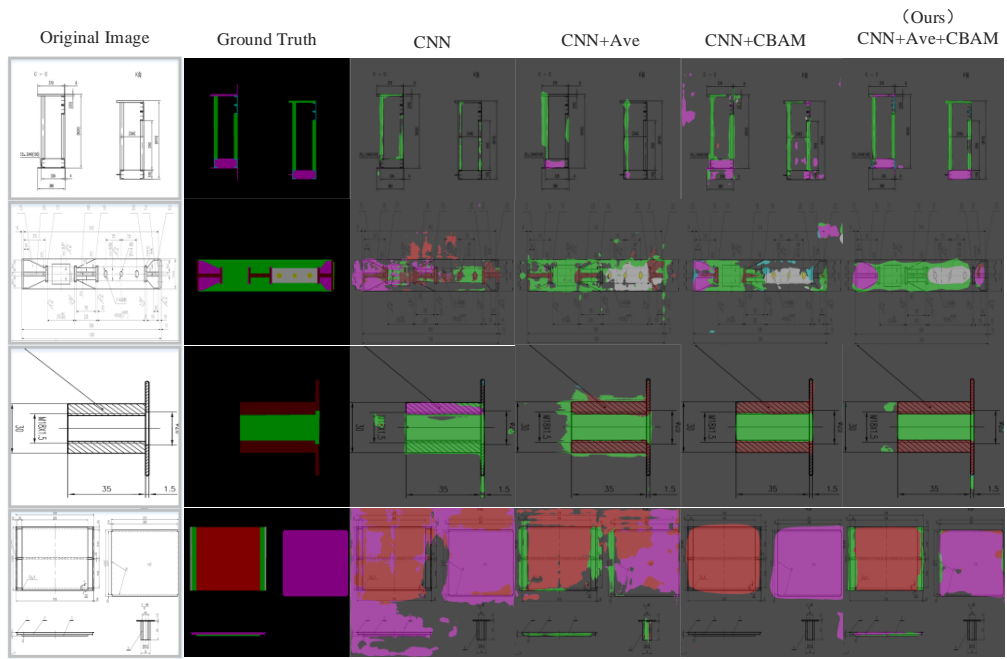

Fig 10. Visualization of segmentation results for successive convolution operations. Ave is represented as the average pooling and convolution operation of the green square in Fig 3, and CBAM is the attention module, as shown in the orange area in Fig 3.

Using the method proposed in this paper combined with CNN to carry out the segmentation extraction experiment of the specific contour of the heavy industry welding engineering drawing, its segmentation performance is shown in Table 4. It can be seen from Table 4 that the method proposed in this paper has a good improvement in the segmentation performance of specific contours of heavy industrial equipment graphics.

| Method | IoU | mAP | Accu |
|---|---|---|---|
| CNN（Base） | 0.4143 | 0.5755 | 0.8867 |
| Base+Ave | 0.4633 | 0.6217 | 0.9130 |
| Base+CBAM | 0.5698 | 0.7101 | 0.9377 |
| Base+Ave+CBAM | **0.6509** | **0.7294** | **0.9518** |

Table 4. In the comparison of different methods for the segmentation results of specific welding engineering units, Ave is expressed as the average pooling and convolution operation, and CBAM is the attention mechanism. Same benchmark, bold indicates excellent.

To further verify the scientificity and rigor of the method proposed in this study, CBAM-U-net with Focal loss and U-net with Focal loss were used to classify the multi-type lines in welding engineering graphics. The confusion matrix is used to realize the visualization of multi-type line classification in welding engineering graphics, and the result is shown in Fig 11. Column elements in the confusion matrix represent true label values for different types of lines, and rows represent true predicted values for different types of lines. Use "Thi", "Thin", "Dash", "Arrow" and "Numer" to represent the thick solid line, thin solid line, dashed line, arrow line, and numbered line in the content of the welding engineering drawing, respectively. It can be concluded from the classification visualization results that the improved model CBAM-U-net proposed in this paper has improved prediction performance for different types of lines compared with the traditional U-net. The results are mainly reflected in the prediction effect of the thick solid line, thin solid line, dotted line, and arrow line.

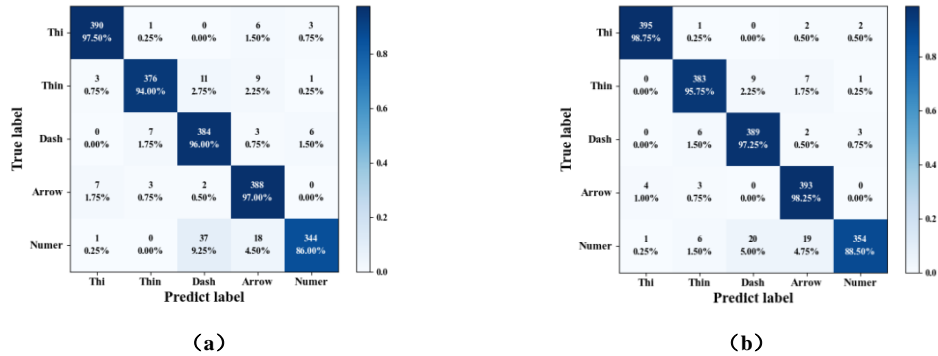

(a)　　　　　　　　　(b)

Fig 11. Comparison of confusion matrix results between U-net and CBAM-U-net models ((a) U-net, (b) CBAM-U-net(Ours)).

3.4 *Results and Discussion*

The results are shown in Table 3. The original U-net model is applied to the segmentation and extraction of specific units of welding structure engineering drawings, and the average accuracy of the intersection-over-union ratio (IoU) can reach 62.62%, and the category means (mAP) and accuracy (Accu) are 0.6775 and 99.37%, respectively. The third column in Fig 8 is the segmentation visualization result of the traditional U-net. It can be seen from the figure that the segmentation accuracy of the traditional U-net in the welding engineering-specific unit has some shortcomings. To solve this problem, this paper proposes to add average pooling + two convolution operations in the horizontal encoder jump structure, and at the same time proposes a vertical jump fusion of high-dimensional vertical encoder feature clusters and low-dimensional horizontal encoder feature clusters-model one. Model one adopts the double-pooling convolution jump structure of the horizontal encoder and the vertical encoder, which reduces the dimensional difference between the encoder and the decoder based on improving the fusion of global semantic features, and realizes the improvement of its segmentation performance. The intersection-over-union ratio (IoU) accuracy of model one segmentation to extract specific units of welding engineering graphics is 63.24%, and the category means (mAP) and accuracy (Accu) are 0.6947 and 96.05%, respectively. However, as can be seen from Fig 8, model one still performs poorly. Immediately, an improved model of adding an attention mechanism to the traditional U-net jump structure was proposed-model two. Model two pays more attention to the channel and spatial information of global semantic features, and reduces the difference of global feature information to achieve better segmentation. The intersection-over-union ratio (IoU) accuracy of model two segmentation to extract specific units of welding engineering graphics is 72.99%, and the category means (mAP) and accuracy (Accu) are 0.7549 and 97.45%, respectively. To make the model better extract the global feature information, the improved model of average pooling, vertical jumping, and attention module is added to the traditional U-net horizontal jumping structure at the same time—Model three. The improved model three can not only reduce the global information ambiguity but also reduce the information dimension difference between the encoder and the decoder and greatly improve its segmentation performance as a whole. The intersection-over-union ratio (IoU) of the specific unit of welding engineering graphics extracted by model three segmentation is 84.72%, and the category means (mAP) and accuracy (Accu) are 0.8684 and 99.42%, respectively. All in all, the performance of the three improved models proposed in this paper for segmenting specific units of welding engineering graphics has been significantly improved compared with the traditional U-net. Among them, the IoU, mAP, and Accu of model three are improved by 22.10%, 19.09%, and 0.05% respectively compared with the traditional U-net in the segmentation task of the specific unit of the welding pattern. To prove the scientificity and rigor of the method proposed in this study, a series of multiple 3x3 convolutional networks were used to continue further verification. From the experimental results in Fig 10 and Table 4, it can be concluded that the method in this paper can effectively improve the specific unit segmentation performance of its model for welded structure engineering graphics.

This paper discusses the application of the U-net-based improved model to the segmentation and extraction task of specific units in welding engineering graphics. It mainly realizes the automatic cutting of sheet metal parts through artificial intelligence machine vision and improves the manufacturing efficiency of heavy industry equipment. The content of this section mainly describes how to improve the segmentation model's attention to global semantic feature information and reduce the dimensional difference between shallow encoder features and deep decoder features. That is, based on adding the attention mechanism, the cross-bar coder jump fusion + convolution operation is adopted. During the experiment, our improved model segmented and extracted specific units with higher performance. However, this method is currently only suitable for the cutting of welded parts in the manufacturing process of large-scale heavy equipment, and may not meet the needs of high-precision mechanical manufacturing processes. Still, it's a good place to start when it comes to mechanical sheet metal automatic cutting tasks.

## 4. Conclusion

Deep learning image segmentation technology has achieved excellent performance in many engineering fields. In this study, it is proposed to use the U-net network to realize the segmentation and extraction of specific units of welding structural engineering graphics in heavy industrial equipment manufacturing, so that the cutting device can automatically cut sheet metal parts by machine vision, thereby improving manufacturing efficiency. Based on the research and analysis of the existing U-net improved model, we propose to add the CBAM module and design the double pooling jump model structure of the upper and lower encoders to realize the global semantic fusion of image features. Not only that, this paper performs two convolution operations on the semantic feature clusters after double pooling to reduce the dimensional difference between the encoder and decoder. The proposed method is trained and validated on a dataset of engineering graphics of complex welded structures. Experimental results show that the performance of our proposed improved model for segmenting specific units of welded structural engineering graphics is better than that of traditional U-net.

**Acknowledgment**